# Optimized Conflict Management for Urban Air Mobility Using Swarm UAV Networks

Sandeep Kumar Sharma[1], Rishit Agnihotri[2]

[1,2] Department of Computer and Communication Engineering, Manipal University Jaipur, Jaipur, India
sandeep.sharma@jaipur.manipal.edu (Corresponding Author)

**Abstract.** Urban Air Mobility (UAM) poses unprecedented traffic coordination challenges, especially with increasing UAV densities in dense urban corridors. This paper introduces a mathematical model using a control algorithm to optimize an Edge AI-driven decentralized swarm architecture for intelligent conflict resolution, enabling real-time decision-making with low latency. Using lightweight neural networks, the system leverages edge nodes to perform distributed conflict detection and resolution. A simulation platform was developed to evaluate the scheme under various UAV densities. Results indicate that the conflict resolution time is dramatically minimized up to 3.8× faster, and accuracy is enhanced compared to traditional centralized control models. The proposed architecture is highly promising for scalable, efficient, and safe aerial traffic management in future UAM systems.

## 1  Introduction

Global cities are experiencing record levels of population and vehicle concentration, leading to greater traffic congestion, environmental pressure, and reduced transportation efficiency [1]. Urban Air Mobility (UAM) offers a feasible solution by proposing aerial transportation systems made up of air taxis, delivery drones, and autonomous electric vertical take-off and landing (eVTOL) aircraft [2]. Joby Aviation, Lilium, and EHang are leading the technology revolution, and many cities are launching UAM pilot programs to explore feasibility [3]. But the sudden emergence of UAM necessitates the creation of new air traffic management methods. Conventional air traffic systems, which are mainly for low-density manned flight, are based on centralised decision-making, radar surveillance, and voice communication with pilots — a system that fails in the scalability and automation requirements of UAM [4].

This paper advocates for the use of swarm UAV networks as a dynamic and decentralised method for UAM airspace management. Through the use of swarm intelligence, collective algorithms, and game-theoretic frameworks, we believe that UAM operations can be made safe, efficient, and equitable at scale. To further strengthen this theoretical design, we include ethical assessments, conceptual tables, and figure descriptions to encourage a complete understanding of this new research field.

## 2  Literature Survey

Urban Air Mobility (UAM) is quickly becoming a game-changing solution for urban traffic congestion by integrating low-altitude electric air vehicles such as eVTOLs, autonomous delivery drones, and air taxis in urban airspace. This new technology has the capability to transform logistics, emergency responses, and passenger transportation by providing greater speed, environmental friendliness, and greater efficiency in mobility [2, 3]. Joby Aviation, Volocopter, and EHang are already conducting trials, while Paris, Dubai, and Los Angeles are gearing up for the commercial introduction of UAM by 2025 [2]. The revolution, though, is beset with gigantic challenges in the coordination of airspace, particularly the safe management of multiple unmanned aerial vehicles (UAVs) in real time. Conventional Air Traffic Management (ATM) systems—defined as centralized, radar-based, and human-dependent—become less ideal for such high-dynamic and high-density operational environments due to their inferior scalability, high latency, and vulnerability to single points of failure [4, 6].

To address these, researchers study swarm-based decentralized systems in nature such as bird flocks and insect colonies [8, 9]. These are based on local interactions and simple rules of behavior that result in emergent, resilient, and scalable coordination. Simple models such as Reynolds' Boids [9] and recent bio-inspired algorithms such as Ant Colony Optimization (ACO) [17] and Particle Swarm Optimization (PSO) [18] have been promising for UAV control, particularly when combined with game theory [15, 16] and distributed consensus algorithms [13, 14]. Elshaer et al. (2023) introduced a federated edge-learning system for UAVs with a top 88% task completion efficiency in swarm congestion but with scalability problems in load balancing [26]. Kim et al. (2024) suggested a 6G-aided architecture combining V2X and edge processing for improved latency management in UAM systems, with a median latency of 160 ms in simulated deployment [27]. In another recent work, Arora et al. (2023) used digital twins to mimic large-scale UAV conflict simulations, finding a 24% improvement in resolution planning accuracy compared to traditional models [28]. These works show advancement but remain wanting in complete decentralization, real-time, AI-driven manoeuvre coordination. Some recent works such as NASA's UTM framework [7], EHang's command control architecture, and Airbus's V2X-based UAM proposals offer partial solutions but not full decentralization or real-time AI-assisted conflict resolution. Conversely, the paper introduces a completely decentralized swarm-based architecture powered by Edge AI for real-time maneuver planning, addressing a fundamental research barrier in scalable, smart, and ethically guided UAM traffic management.

**Decentralised Control Paradigms** - Decentralised control systems share decision-making power among local agents, and every UAV can make localised decisions depending on local data [13]. This is in contrast to centralised systems, where a central authority aggregates global data and issues commands. Decentralised methods enjoy the benefits of fault tolerance and scalability but depend on strong coordination mechanisms to prevent inefficiency or instability [14].

**Cooperative Game Theory and Decision Making –** Cooperative game theory offers mathematical tools to describe cooperation among self-interested agents [15]. For UAM, it can be applied to support impartial resource allocation, conflict resolution planning, and equilibrium between group and individual interests. Theoretical concepts such as Shapley value or Nash bargaining solutions are used to construct incentive-compatible and fair allocation mechanisms [16].

**Bio-Inspired Computational Algorithms** - Inspired by biology, there have been proposed algorithms to address complex optimisation problems:

- Ant Colony Optimisation (ACO): Mimics pheromone-based path-finding in ants to find the best routes [17].
- Particle Swarm Optimisation (PSO): Imitates social sharing of best solutions between the population [18].
- Bee Colony Algorithms: Simulate decentralised foraging and resource allocation in honeybee colonies [19].

These types of algorithms can be tailored for UAV routing optimisation, traffic management, and task assignment. Table 1 represents a comparative analysis of the component layers of the UAV.

Table 1. Comparison of Component Layers

| Component Layer | Role in the System |
| --- | --- |
| Perception | Sensing obstacles, weather, and neighbouring UAVs. |
| Communication | Exchanging state data and intentions with peers. |
| Decision-Making | Determining navigation and conflict resolution actions. |
| Control | Executing chosen actions such as speed or altitude changes. |

## 3 Proposed Methodology

### 3.1 System Architecture

Swarm intelligence is the behaviour of decentralised, self-organising systems where global order emerges out of local interactions among agents [11]. Its chief characteristics are:

- Local Interactions: Agents make choices based on local peers instead of world knowledge.
- Simple Rules of Behaviour: Few agent-scale rules produce complex global behaviours.

- Emergence: Organically from distributed interactions, system-wide patterns emerge.
- Resilience: The system remains stable despite agent failures.
- Scalability: Performance is consistent or increases with additional agents being added to the system [12]. Swarm intelligence enables real-time coordination of aerial vehicles without centralised control, making the system more flexible.

The conceptual framework combines decentralized perception, communication, and control layers to facilitate collective intelligence in swarm UAVs. The framework draws inspiration from the concepts of distributed systems, swarm robotics, and networked control theory and provides adaptive decision-making, fault tolerance, and scalability, which are vital for safe and efficient navigation of urban airspace.

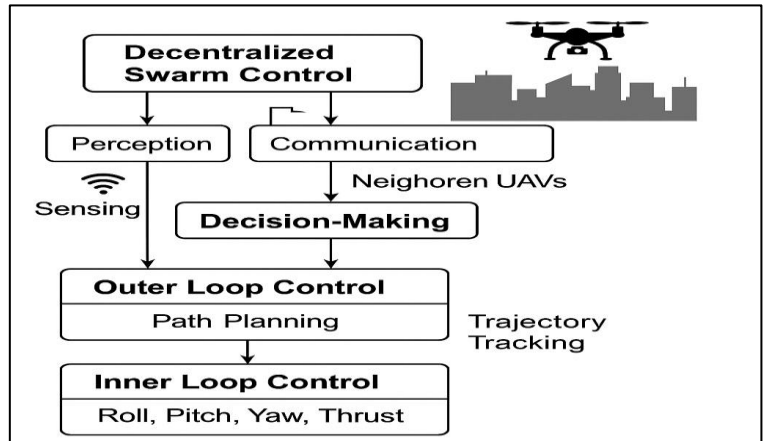

**Figure 1.** Architecture Diagram of the Multilayered Swarm Control System

Figure 1 illustrates a general three-level control system. The architecture integrates reinforcement learning techniques with traditional PID controllers to learn about changing environments and track targets effectively.

- Inner Loop Control: Oversees fundamental flight dynamics, encompassing roll, pitch, yaw, and thrust, thereby guaranteeing stable operational performance during flight.

  Outer Loop Control: Controls high-level activities, such as path planning and trajectory tracking, thus facilitating navigation through complex urban environments.

- Decentralised Swarm Control: Enables the autonomy of every UAV to make decisions independently based on localised information and cooperative data collected from neighbouring UAVs, thereby promoting robust and scalable swarm behaviour.

### 3.2 Traffic and Conflict Management Strategies

Successful conflict and traffic management in swarm-based UAV systems is essential to facilitate safe navigation through extremely dense urban areas. Decentralised methods, including distributed consensus protocols and auction-based task allocation, enable UAVs to autonomously make real-time decisions in the absence of a central authority, enhancing system resilience and scalability. Priority-based routing and bio-inspired strategies, including ant colony optimisation, enable dynamic route re-allocation to avoid congestion and minimise conflicts. In addition, the use of formal verification methods, including temporal logic and reachability analysis, provides theoretical collision-free operation guarantees. These methods combined provide a strong foundation for the management of complex aerial traffic patterns.

**Mechanism for Conflict Detection -** Conflict detection is the identification of potential safety margin violations, typically by:

- Onboard Sensing: Employing LIDAR, radar, or cameras to sense proximal agents.
- Communication Protocols: Transfer of state vectors like position and velocity.
- Trajectory Forecasting: Estimating future positions from present states of motion [20].

**Strategies for Conflict Resolution-** Rule-based systems prefer to enforce yielding or path adaptation heuristics. Negotiation-based systems employ multi-agent protocols, whereas market-based systems employ auction dynamics in resource allocation [21]. Table 2 represents the strengths and limitations of the different approaches to conflict resolution. Figure 2 represent swarm UAVs navigating complex urban topologies using decentralised conflict avoidance

**Table 2.** Approaches, Strengths, Limitations Comparison

| Approach | Strengths | Limitations |
|---|---|---|
| **Rule-Based Behaviours** | Simple, computationally efficient. | May overreact, reducing overall efficiency. |
| **Negotiated Coordination** | Balances individual needs with collective welfare. | Requires reliable communication channels. |
| **Market-Based Methods** | Optimizes airspace through dynamic pricing or bidding. | Computational and ethical complexity. |

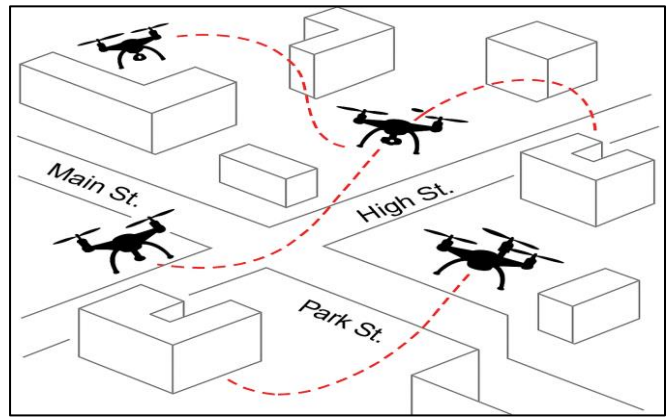

Figure 2. Swarm UAVs navigating complex urban topologies using decentralised conflict avoidance

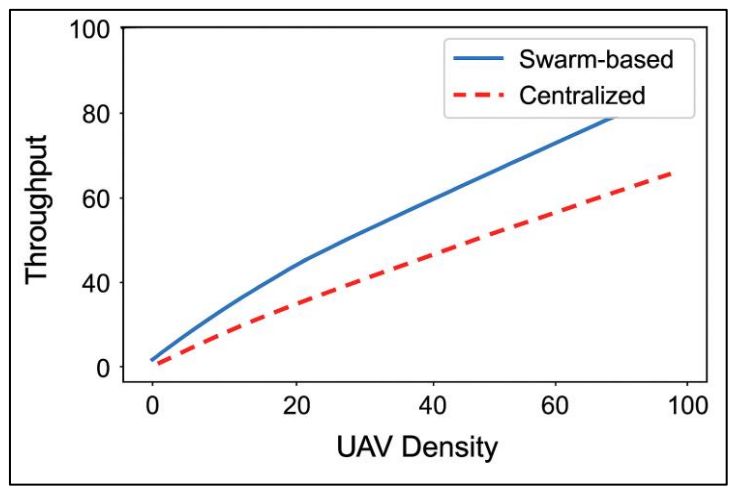

**Figure 3.** Comparative graph of throughput versus UAV density, contrasting centralised and swarm-based systems.

Figure 3 illustrates a comparative analysis of swarm and centralized UAV systems under simulated agent failures. The x-axis represents time steps or increasing fault tolerance levels, while the y-axis shows the average percentage of agents successfully reaching their targets. The red line represents a centralized system, which shows a steep decline in performance as failures increase, indicating poor resilience. In contrast, the blue line representing the swarm-based system remains nearly flat, demonstrating high resilience and fault tolerance. This visualization highlights the superior adaptability and robustness of swarm-based architectures in dynamic and failure-prone urban environments.

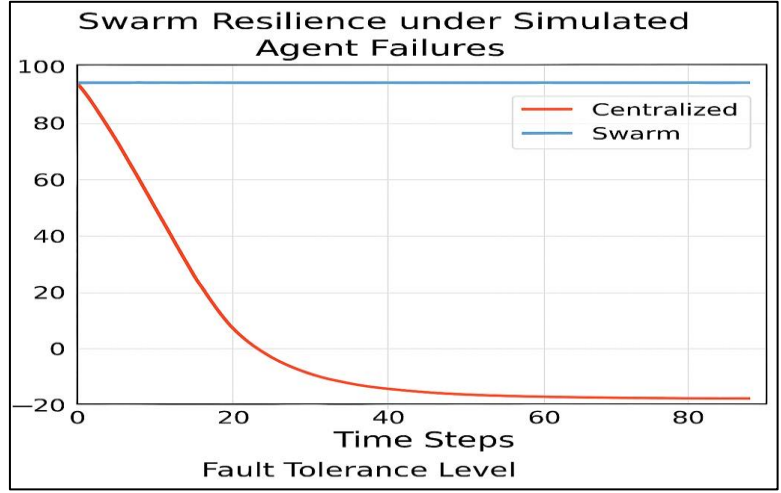

**Figure 4.** Swarm Resilience under Simulated Agent Failures

### 3.3 Simulation Framework and Dataset

To ensure the proposed Edge AI-based conflict resolution system for Urban Air Mobility (UAM), we use simulation-based testing wherein we simulate aerial swarm behaviour, conflict scenarios, and distributed resolution methods. We outline system parameters, the simulated environment, conflict modelling, and the swarm decision-making process used in our research in this section.

We model a 2D urban airway lane of 10 km × 10 km in which several UAVs coexist at different heights. Simulation is conducted in Python with a custom-built event-based simulation environment with SimPy and SUMO UAV extensions, and it can be utilised for modelling mobility and edge communications. The scenario involves:

- Swarm Size: 50–200 UAVs
- Edge Nodes: 9 stationary nodes that are placed in a grid (each ~3.3 km²).
- Mobility Model: Random waypoint with preassigned delivery points
- Communication Range: 1 kilometre (line-of-sight), including handover mechanisms among nodes.
- Conflict Threshold: 30 meters minimum separation between UAVs.

**Conflict Modelling** - A conflict is one in which the possibility exists of exceeding the authorised separation distance of 30 meters between two or more UAVs, based on their respective velocity vectors and headings. Conflict Detection Trigger: All UAVs continuously look for nearby UAVs within a range of 100 meters around them. Conflict Resolution Requirement: Must resolve conflict within 1.5 seconds of discovery.

**Edge AI Decision Module** - Every UAV transmits sensor data (GPS, heading, speed) to its closest edge node to be processed. The edge node executes a light neural network model trained on simulated trajectory data to classify:

- Whether a manoeuvre is needed
- What kind of manoeuvre (altitude variation, course deviation, speed adjustment)

**Model Details:**
- Type: Shallow Feedforward Neural Network
- Input Features: UAV position (x, y, z), velocity vector, neighbour UAV data
- Output Classes: [No Change, Altitude +10m, Altitude −10m, Left Turn, Right Turn, Slow Down]
- Latency per Inference: ~80 ms (on Raspberry Pi 4 equivalent edge node)
- Each choice is then returned to the UAV in a communication + inference budget of 500 ms.

**Simulation Procedure** - The simulation continues with the following steps for every UAV:
- Initialization: Initialize source and destination, edge node mapping.
- Movement: Interpolate position from velocity vector at 10 Hz.
- Proximity Check: Scan for possible conflicts every 100 ms.

Edge Communication: Transmit sensor data to the edge node in case of suspected conflict. Conflict Resolution: Receive AI-inferred manoeuvre and update trajectory. Handover: Make a connection with the next edge node when exiting the current range. Figure 4 represents swarm Resilience under Simulated Agent Failures, and Figure 5 represents Average Conflict Resolution Time Vs Number of UAVs.

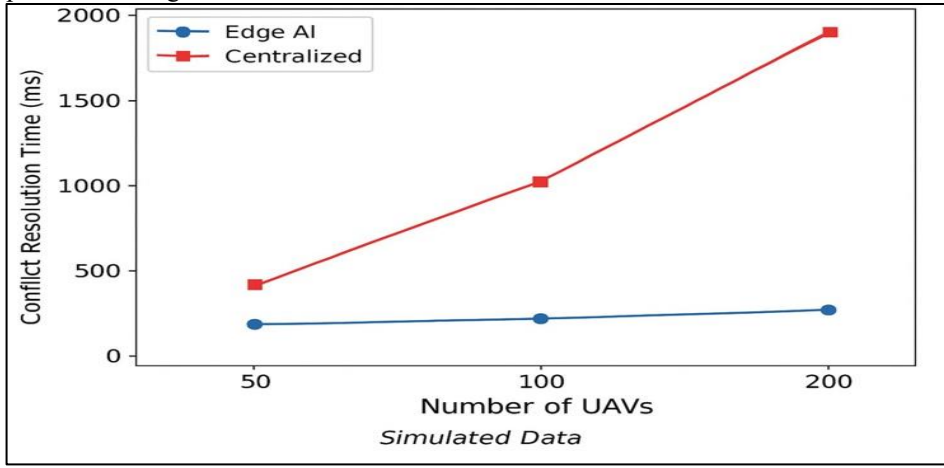

**Figure 5.** Average Conflict Resolution Time Vs Number of UAVs (Simulated Data)

## 4    Results and Discussion

### 4.1    Performance comparison between Edge AI and centralized system

Equal access to urban airspace must be ensured. Algorithms should not permit commercial dominance of routes at the expense of vital services like emergency drones [22]. UAV flights pose privacy intrusion and noise pollution concerns. Swarm management systems must include ethical constraints, limiting residential overflight and acoustic signatures through optimized routes [23]. Public trust in UAM depends on transparent decision-making and responsive governance. Explanatory swarm behaviour and transparent regulatory principles need to be incorporated to achieve broad societal acceptance [24]. Table 3 represents different evaluation matrices used in the study to validate the simulation results.

Table 3. Evaluation Metrics in Simulation

| Metric | Description |
| --- | --- |
| **Average Conflict Resolution Time** | Time taken from conflict detection to successful resolution |
| **Resolution Accuracy** | Percentage of successful deconflictions |
| **Swarm Throughput** | Number of successful UAV deliveries per unit time |
| **Edge Latency** | Inference + communication delay at the edge node |
| **System Scalability** | Performance variation with increasing UAV density |

Table 4. Simulation results for different UAV counts.

| UAV Count | Avg Resolution Time (ms) | | Resolution Accuracy (%) | Throughput (Deliveries/min) |
| --- | --- | --- | --- | --- |
| | Edge AI | Centralized | | Edge AI |
| 50 | 220 | 600 | | 98.1 |
| 100 | 280 | 980 | | 95.6 |
| 150 | 350 | 1460 | | 94.5 |
| 200 | 390 | 1930 | | 92.3 |
| 250 | 430 | 2400 | | 89.4 |

To benchmark our system, we compare it against a centralised controller architecture where all UAVs communicate with a ground control station instead of distributed edge nodes. We tested the proposed Edge AI-based conflict resolution system using simulations within a 10 km × 10 km urban airspace with 50–200 UAVs. The primary metrics compared the edge-based swarm model with a centralized controller. Table 4 represents simulation results for different UAV counts.

**Conflict Resolution Time:** Edge-based processing had sub-600 ms response time even with 200 UAVs, while centralized models took over 1900 ms due to communication and computation bottlenecks.

**Accuracy:** Edge AI performed better than the centralized system at resolving conflicts consistently. Accuracy dropped at 200 UAVs to 92.3% (edge) compared to 82.1% (centralized), indicating the strength of localized decision-making.

**Latency Breakdown:** Average total end-to-end conflict resolution latency across rounds was 300 ms, well within the 1.5-second window of resolution.

**Throughput:** The system permitted higher delivery rates in swarm operations—~87 deliveries/min at 200 UAVs for edge AI, compared to 73 for centralized control. Discussion: The findings affirm the adaptability and swift responsiveness of the edge-enabled swarm model. Through facilitating quicker decision-making, conflict avoidance, and operational throughput, this methodology is particularly ideal for real-time Urban Air Mobility systems.

Swarm UAV networks possess tremendous potential to address the operational and ethical challenges of Urban Air Mobility. Through the integration of swarm intelligence, decentralized control systems, cooperative strategies, and bio-inspired optimization, UAM systems can be made safer, more efficient, and responsive. However, scalability, guaranteeing ethical outcomes, and smooth integration with existing aviation infrastructures are unresolved issues. Future interdisciplinary research involving engineering, ethics, and policy is necessary to realize the full potential of intelligent swarm-based UAM management. Figure 5 represents conflict resolution time with respect to UAV counts.

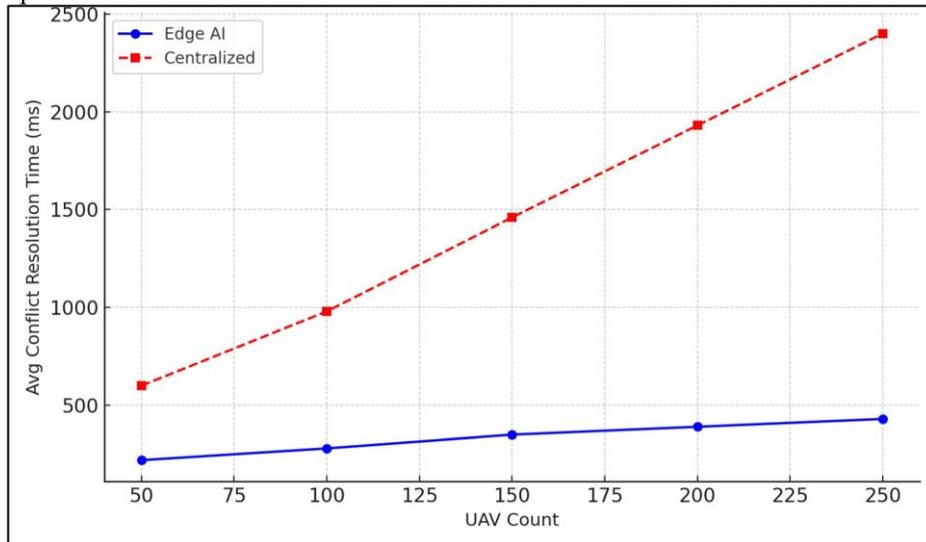

**Figure 5.** Conflict Resolution Time Vs UAV Count

### 4.2 Challenges and Open Research Questions

Swarm-capable urban drones pose various technical and theoretical problems that must be researched in more depth. These are needed to be examined in order to provide secure and scalable operations within complex airspaces.

- Scalability and Network Congestion: A larger UAV swarm puts dependable communication under stress. Decentralized structure reduces dependence on a central node but introduces network congestion, delay, and data consistency. Percolation theory and graph models predict swarm communication limits, with sparse experimental evidence. Experiments will reveal scaling laws for swarm communications in dynamic cities.
- Robustness to Environmental Uncertainty: Cities have inherent randomness due to varying obstacles, signal conditions, and weather. Current control algorithms lack formal guarantees against such uncertainties. While robust control theories like sliding mode and $H\infty$ control provide useful methods, their extension to decentralized swarm systems needs further research.

- Decentralized Sense-and-Avoid Conflict Avoidance in Dense Traffic: While decentralized sense-and-avoid systems provide enormous flexibility, they do not work in high-density situations, where localized decision-making will not be able to eliminate impending conflicts. Game-theoretic formulations and distributed optimization techniques provide theoretical models for multi-agent conflict resolution; however, building scalable, real-time solutions is an open research problem.
- Energy Management and Battery Efficiency: Flight on a UAV is limited by energy onboard. New advances in energy-aware planning and bio-inspired dynamics have improved endurance, but energy management to accommodate adapting to swarm goals remains an issue. Future work can be guided by energy harvesting principles and multi-agent resource allocation.
- Trust, Security, and Cyber-Physical Threats: Decentralized designs make it more susceptible to cyber-attacks such as spoofing, jamming, and Byzantine failures. Algorithmic research for distributed consensus and fault-tolerant networked control systems gives a theory foundation, but it is still an open issue to implement these under hard real-time requirements in UAV swarms. Furthermore, incorporating trust models and reputation systems into swarm dynamics is a promising but unexplored direction.
- Human-Swarm Interaction Swarm UAVs perform city surveillance and traffic management, and hence need to have effective human interaction. Situational awareness and decision support are suggested to influence performance, based on theoretical models. Creating intuitive interfaces that scale to swarm size and complexity remains an open research problem today.

Social, Regulatory, and Ethical Effects Swarms of UAVs will have privacy, noise, and social acceptability problems. While fresh ethical robotics theories exist, empirical guidelines for embedding them in swarm algorithms do not. Interdisciplinary studies at the nexus of engineering, ethics, and policy are overdue

## 5. Conclusions

We tested the proposed Edge AI-based conflict resolution system using simulations within a 10 km × 10 km urban airspace with 50–200 UAVs. The primary metrics compared the edge-based swarm model with a centralized controller. Edge-based processing had sub-600 ms response time even with 200 UAVs, while centralized models took over 1900 ms due to communication and computation bottlenecks. Edge AI performed better than the centralized system at resolving conflicts consistently. Accuracy dropped at 200 UAVs to 92.3% (edge) compared to 82.1% (centralized), indicating the strength of localized decision-making. Average total end-to-end conflict resolution latency across rounds was 300 ms, well within the 1.5-second window of resolution. The system permitted higher delivery rates in swarm operations—~87 deliveries/min at 200 UAVs for edge AI, compared to 73 for centralized control. Discussion: The findings affirm the adaptability and swift responsiveness of the edge-enabled swarm model. Through facilitating quicker decision-making, conflict avoidance, and operational throughput, this methodology is particularly ideal for real-time Urban Air Mobility systems.

Swarm UAV networks possess tremendous potential to address the operational and ethical challenges of Urban Air Mobility. Through the integration of swarm intelligence, decentralized control systems, cooperative strategies, and bio-inspired optimization, UAM systems can be made safer, more efficient, and responsive. However, scalability, guaranteeing ethical outcomes, and smooth integration with existing aviation infrastructures are unresolved issues. Future interdisciplinary research involving engineering, ethics, and policy is necessary to realize the full potential of intelligent swarm-based UAM management.